\documentclass[conference]{IEEEtran}
\IEEEoverridecommandlockouts
\usepackage{cite}
\usepackage{amsmath,amssymb,amsfonts}
\usepackage{algorithmic}
\usepackage{graphicx}
\usepackage{textcomp}
\usepackage{xcolor}
\usepackage{algorithm}
\usepackage{algorithmic}
\usepackage{booktabs}
\usepackage{multirow}
\usepackage{hyperref}

\newcommand{\tabincell}[2]{\begin{tabular}{@{}#1@{}}#2\end{tabular}}
\def\BibTeX{{\rm B\kern-.05em{\sc i\kern-.025em b}\kern-.08em
		T\kern-.1667em\lower.7ex\hbox{E}\kern-.125emX}}

\usepackage[]{caption}

\begin{document}
	\title{PFL-MoE: Personalized Federated Learning Based on Mixture of Experts
	}
	
	\author{
		\IEEEauthorblockN{1\textsuperscript{st} Binbin Guo}
		\IEEEauthorblockA{\textit{School of Comp. Sci. \& Eng.} \\
			\textit{Sun Yat-sen University}\\
			Guang Zhou, China \\
			guobb5@mail2.sysu.edu.cn}
		\\
		\IEEEauthorblockN{4\textsuperscript{th}  Weigang Wu}
		\IEEEauthorblockA{\textit{School of Comp. Sci. \& Eng.} \\
			\textit{Sun Yat-sen University}\\
			Guang Zhou, China \\
			wuweig@mail.sysu.edu.cn}
		\and
		\IEEEauthorblockN{2\textsuperscript{nd} Yuan Mei}
		\IEEEauthorblockA{\textit{School of Comp. Sci. \& Eng.} \\
			\textit{Sun Yat-sen University}\\
			Guang Zhou, China \\
			meiy7@mail2.sysu.edu.cn}
		\\
		\IEEEauthorblockN{5\textsuperscript{th}Ye Yin}
		\IEEEauthorblockA{\textit{Interactive Entertainment Group} \\
			\textit{Tencent Inc.}\\
			Shenzhen, China \\
			dbyin@tencent.com}
		\and
		
		\IEEEauthorblockN{3\textsuperscript{rd}Danyang Xiao }
		\IEEEauthorblockA{\textit{School of Comp. Sci. \& Eng.} \\
			\textit{Sun Yat-sen University}\\
			Guang Zhou, China \\
			xiaody@mail2.sysu.edu.cn}
		\\
		
		\IEEEauthorblockN{6\textsuperscript{th}Hongli Chang}
		\IEEEauthorblockA{\textit{Interactive Entertainment Group} \\
			\textit{Tencent Inc.}\\
			Shenzhen, China \\
			honglichang@tencent.com}
	}
	\maketitle

	\begin{abstract}
		Federated learning (FL) is an emerging distributed machine learning paradigm that avoids data sharing among training nodes so as to protect data privacy. Under coordination of the FL server, each client conducts model training using its own computing resource and private data set. The global model can be created by aggregating the training results of clients. To cope with highly non-IID data distributions, personalized federated learning (PFL) has been proposed to improve overall performance by allowing each client to learn a personalized model. 
		However, one major drawback of a personalized model is the loss of generalization.
		To achieve model personalization while maintaining generalization, in this paper, we propose a new approach, named PFL-MoE, which mixes outputs of the personalized model and global model via the MoE architecture.
		PFL-MoE is a generic approach and can be instantiated by integrating existing PFL algorithms. Particularly, we propose the PFL-MF algorithm which is an instance of PFL-MoE based on the freeze-base PFL algorithm. We further improve PFL-MF by enhancing the decision-making ability of MoE gating network and propose a variant algorithm PFL-MFE. We demonstrate the effectiveness of PFL-MoE by training the LeNet-5 and VGG-16 models on the Fashion-MNIST and CIFAR-10 datasets with non-IID partitions.
	\end{abstract}
	
	
	\section{Introduction}
	In the past few years, we have witnessed the rapid development of deep learning in artificial intelligence (AI) applications, such as computer vision, natural language processing, and recommendation systems.
	As is known to all, the success of deep learning technologies relies on massive data set which are usually the collections of data from different organizations, devices, or users. 
	And distributed machine learning (DML) \cite{ben2019demystifying}, which may employ a large number of worker nodes to collaboratively conduct training, has been widely used nowadays to train large scale neural network models on large volume data set. 
	However, in many scenarios, data is in fact distributed among different clients (owned by different users) and is sensitive to privacy. Such data cannot be collected and shared with any other nodes. 
	On the other hand, as the storage and computational capabilities of edge devices grow, it is becoming increasingly attractive to perform learning directly on the edge \cite{2019In}.
	Unfortunately, it is difficult for a client to learn an effective model using only its own data, because the data set at a single client is usually not large enough. 
	Federated Learning (FL) allows a large number of clients (e.g., mobile phones and IoT devices) to learn a global model together without data sharing. The global model is created by repeatedly averaging model updates from small subsets of clients. 
	More specifically, each round of FL includes three basic phases: (1) the FL server delivers the global model to each participating client; (2) the clients conduct training based on its private data and return the updated models to the server; and (3) the FL server gets the latest global model by aggregating the uploaded models. 
	As the training converges, the final global model can achieve performance similar to the model trained via pooled and shared data.
	
	One of major challenges of FL is the statistical diversity among data owned by different clients. 
	The distribution of data across clients is natural inconsistency and non-IID (IID, independent and identically distributed), and the amount of data is also highly biased.
	The non-IID characteristic may largely affect the convergence behaviors of the global model trained, because non-IID data may cause the model parameters fluctuates largely and delay or even destroy the convergence. 
	It will be even worse if privacy-preserving mechanisms, e.g., differential privacy \cite{46432}, are taken into consideration.
	Besides, different clients have great accuracy differences under the same global model. In some clients, the global model is even worse than the locally trained model.
	
	Personalization of the global model is necessary to handle the statistic heterogeneity of the distributed data.
	In personalized federated learning (PFL), each client learns a personalized model suitable for its own data distribution, which is more flexible than a single global model.
	Yu et al. \cite{yu2020salvaging} has investigated various technologies for adapting the federated global model to an individual client, including fine-tuning \cite{wang2019federated}, multi-task learning \cite{smith2017federated}, and knowledge distillation \cite{li2019fedmd}. 
	These PFL algorithms can effectively improve the model performance for individual participants. Almost all personalized models outperform the corresponding locally trained models, which greatly motivates clients to participate in federated learning.
	However, personalization is always accompanied by the loss of generalization, which is manifested in the increases of generalization error of personalized model. 
	For example, a client has only a small amount of data, and the personalized model is likely to overfit the local data.
	How to balance personalization and generalization is the current research focus of personalized federated learning.
	
	
	Most of existing PFL algorithms attempt to find a single model that is both personalized and generalized.
	Alternatively, FL+DE \cite{peterson2019private} combines two models to improve accuracy based on the mixture of experts (MoE) \cite{jacobs1991adaptive}.
	%
	MoE is one of the most popular combining methods. It is established based on the divide-and-conquer principle in which the problem space is divided between a few neural network experts, supervised by a gating network \cite{2014Mixture}.
	In FL+DE, all clients collaborate to build a general global model, but maintain private, domain-adapted expert models. In each client, the global model and the private model are regarded as global expert and domain expert respectively and their outputs are mixed by the gating network. 
	The predictions of the two models are much more accurate than those of the single global model.
	
	
	
	Inspired by FL+DE, we propose a personalized federated learning approach PFL-MoE that considers the personalized model as local expert and combines it with the federated global model through MoE architecture. 
	PFL-MoE is a generic approach and can be instantiated by existing PFL algorithms.
	Particularly, we propose PFL-MF algorithm, which is an instance of PFL-MoE based on the freeze-based PFL (PFL-FB) algorithm \cite{yu2020salvaging}. 
	PFL-FB is an effective algorithm  adapting the federated global model to an individual client.
	
	Moreover, we find that the MoE architecture cannot work well with complex input data, such as the color images, because it is difficult for the gating network to extract useful features from high-dimensional input data for decision-making.
	To enhance the decision-making ability of MoE gating network, we modify the model structure of PFL-MF and propose an extended algorithm PFL-MFE.
	More precisely, PFL-MFE uses abstract features extracted by the base layer as the input of MoE gating network.
	
	We conduct extensive experiments to evaluate the performance of our algorithms and compare them with existing ones. We use three combinations of dataset and model: Fashion-MNIST + LeNet-5, CIFAR10 + LeNet-5, and CIFAR-10 + VGG-16. In various non-IID experimental settings, the proposed algorithms achieve higher global accuracy while the same or even better local test accuracy than PFL-FB. 
	The results show our approach combines the advantages of the personalized model and the global model.

	\textbf{Contributions:}
	\begin{itemize}
		\item We propose a novel PFL approach, namely, PFL-MoE, that can achieve both personalization and generalization. PFL-MoE is generic and can be instantiated using existing FL algorithms. 
		\item Based on our PFL-MoE approach, we propose the PFL-MF algorithm and its extension, the PFL-MFE algorithm. The former is an instance of PFL-MoE and the later improves PFL-MF by modifying the MoE architecture so as to achieve better decision-making ability.
		\item We conduct experiments to evaluate our approach. An overall evaluation of PFL-MF and PFL-MFE algorithms is provided. The experiment results confirm the advantage of our design. 
		
	\end{itemize}
	
	The rest of the paper is organized as follows. 
	Section \ref{sec:relatedworks} reviews related works on personalization techniques for federated learning. Section \ref{sec:pflmoe} describes the detailed design of the PFL-MoE approach, and the instance algorithms PFL-MF and PFL-MFE are presented in Section \ref{sec:pflmf}. Section \ref{sec:dbe} reports the performance evaluation results. Finally, Section \ref{sec:conclusion} concludes the paper.

	\section{Related Works} \label{sec:relatedworks}
	
	\subsection{Classical Federated Learning}
	The first FL system is developed by Google \cite{mcmahan2017communication,45648}, which selects a sample of available smartphones to update the language prediction model based on the edge server architecture, and has got a good result.
	However, there are still many challenges in federated learning, including privacy concerns, communication cost, systems heterogeneity and statistical heterogeneity \cite{li2020federated, kairouz2019advances}. 
	Although federated learning only communicates model parameters,
	studies have shown that monitoring model updates can reversely derive data, making privacy protection an important issue \cite{melis2019exploiting}.
	A considerable amount of literature has been published on privacy leakage risks and protection methods \cite{lyu2020threats}.
	The FL server coordinates the participants to iterative training a global model, where uploading and downloading model parameters require a lot of communication overhead. Some studies aim to reduce communication costs in FL \cite{45648}.
	In federated settings, all edge devices can be clients. In most cases, these devices have fewer network resources and cannot keep online all the time \cite{45648}.
	Moreover, 
	the software and hardware resources and computing power of each client vary greatly. These issues requires the federated learning framework to coordinate and allocate the clients effectively \cite{DBLP:conf/mlsys/LiSZSTS20}.
	
	\subsection{Handling Non-IID and Personalization}
	Given the distributed nature of the clients, the statistics of the data across them are likely to differ significantly,
	which is another big challenge for FL \cite{kairouz2019advances}. 
	A large number of experiments have evaluated the impact of non-IID on FL \cite{49350}. 
	Simultaneously, various non-IID data construction methods \cite{49350} and the non-IID datasets from the real world \cite{ caldas2018leaf} are emerging.
	Li et al. \cite{DBLP:conf/iclr/LiHYWZ20} provided the first proof of the convergence of FedAvg\cite{mcmahan2017communication}, widely used in federated learning, under non-IID data distributions.
	Zhao et al. \cite{zhao2018federated} proposed a method of sharing part of the global data to alleviate 
	data dissimilarity among the clients, so as to improve the convergence rate and performance of the global model.
	In FedProx \cite{DBLP:conf/mlsys/LiSZSTS20}, the proximal term was added to each client's optimization goal to limit the drift of its update so that the algorithm has better robustness and convergence stability than FedAvg. 
	
	Alternatively, personalized federated learning \cite{kulkarni2020survey} starts from the incentive of the clients participating in federated learning, and turns the goal to improve the accuracy of the personalized models of the clients with non-IID data distributions, which improves the overall performance effectively. A large and growing body of literature has investigated to achieve personalization through various methods:
	\paragraph {Fine-turning and meta-learning}
	In Wang et al.'s work \cite{wang2019federated}, on each client, the personalized model was fine-tuned from the federated global model (PFL-FT), but the used of the model is depends on whether or not it outperforms the global model. In his image classification experiment, most clients can get a better personalized model, and the rest of the clients continue to use the global model. Jiang et al. \cite{jiang2019improving} observed that the FedAvg algorithm can be interpreted as a meta learning algorithm (MAML) \cite{DBLP:conf/icml/FinnAL17}. This provides a good explanation for fine-tuning can achieve good personalization that FedAvg can get a good initial model for the clients which achieve high precision with only a little local training.
	Also based on the MAML algorithm, Fallah et al. proposed a variant of FedAvg named Per-FedAvg \cite{DBLP:conf/icml/FinnAL17} which turns the goal into learning a better initial model for each client and provides theoretical guarantees. 
	Furthermore, pFedMe \cite{t2020personalized} updates both the personalized and global models by solving the bi-level optimization problem instead of using the global model as initialization, which is more flexible than Per-FedAvg.
	
	\paragraph{Multi-task}
	Smith et al. \cite{smith2017federated} observed that multi-task learning is naturally suitable for federated learning. They treated each client as a individual task in multi-task learning literature, and investigated a system-aware approach called MOECHA that takes into account communication costs, straggles, and fault tolerance. Not limited to convex models, Corinzia et al. \cite{corinzia2019variational} proposed the VIRTUAL algorithm to solve generic non-convex models for federated multi-task learning in the scenarios where the data is highly Non-IID. 
	\paragraph{Knowledge distillation}
	A new federated learning framework FedMD \cite{li2019fedmd} was proposed, allowing participants to independently design their own models. FedMD achieves the knowledge transmission between each client through transfer learning and knowledge distillation.
	\paragraph {Mixing models}
	In Hanzely et al.'s work \cite{hanzely2020federated}, a new federated learning optimization goal was defined to seek a mixture of global and local models. The method controls the mixing degree of the local and global models through $\lambda$. When $\lambda=0$, it means each client is trained locally, and when $\lambda=\infty$, it is equivalent to FedAvg algorithm. When taking $\lambda>0$, it means personalization is encouraged. They further theorize about the definition and come up with an algorithmic independent theory.	
	Furthermore, APFL \cite{deng2020adaptive} determines each client's contribution through optimization, with each client contributing to the global model while training its own local model.
	APFL is also a communication-effective algorithm. 
	It is proved that the mixing of the local model and global model can reduce the generalization error.
	Google researchers studied a systematic, individualized learning theory and analyzes three kinds of individualized methods \cite{mansour2020three}, including user clustering, data interpolation, and model interpolation. Model interpolation is obtained from the weighted average of parameter of the local model and global model, and its performance is proved by theory and experiment.
	FL+DE \cite{peterson2019private} uses privacy-preserving FL to train a generic model and mixes it with each user's private domain model using the MoE architecture.
	
	
	\section{PFL-MoE} \label{sec:pflmoe}
	
	PFL-MoE is a generic approach for personalization, which mainly contains the following three stages:
	
	\begin{itemize}
		\item \textit{Federated Learning Stage:} Following the traditional FL framework, each client participates in FL training. 
		The global model is created by repeatedly aggregating model updates from small subsets of clients.
		This stage is agnostic about the aggregation methods or privacy-preserving techniques of federated learning. It only needs the final global model for the next personalization. 
		\item \textit{Personalization Stage:} Each client downloads the latest global model from the FL server, and then conducts local adaption to get a personalized model based on the global model and local data. 
		\item \textit{Mixing Stage:} Via the previous two stages, the client gets two models, one is the personalized model, and the other is the global model. Based on the MoE architecture, the parameters of the gating network are trained to combine the models and make them work together.
	\end{itemize}
	
	\subsection{Federated Learning}
	In federated learning, we consider that the number of clients participating in training is N. We define the model structure of the global model as $M_G$, and define the parameter of the global model as $\theta$ ($d_1$-dimensional vector). At the beginning of training, the global model structure $M_G$ has been determined by the central FL server. The goal of federated learning is to find a $\theta\in \mathbb{R}^{d_1}$ that makes the average 
	expected loss of all clients minimal.
	We define $F: \mathbb{R}^{d_1}\rightarrow\mathbb{R}$ as the global loss of FL, so the learning goal of FL is the following formula:  
	\begin{equation}
		\min_{\theta\in \mathbb{R}^{d_1}}F(\theta) = \frac{1}{N}\sum_{i=1}^{N}f_i(\theta) \label{fed:gloabloss}
	\end{equation}
	Since the data of each client follow a different distribution, we define the data distribution of the client $i$ as $\mathcal{D}_i$. We define the expected loss over $\mathcal{D}_i$ of the client $i$ as $f_i$. In particular, we assume that the learning is a supervised learning task. At this time, $f_i$ can be written as:
	
	\begin{equation}
		f_i(\theta) =\mathbb{E}_{(x,y)\sim\mathcal{D}_i}[L_i(M_G(\theta; x), y)]
	\end{equation}
	where $(x,y)$ is a pair of data composed of the input and the corresponding label, following the distribution $\mathcal{D}_i$. $ L_i $ is the loss function of the client $i$, usually, it is uniform in each client. E.g., in a classification task using a neural network, it is the cross entropy loss function.
	
	\subsection{Personalization}
	

	In PFL, the optimization goal is the same as individual learning. However, in individual learning, due to the insufficient local data, it is difficult for a client to estimate its local data distribution, which leads to a large generalization error.
	Based on the federated global model,
	it is easier for individuals to obtain models with better generalization capabilities than individual learning.
	In particular, we chose PFL-FT, a local adaptation PFL algorithm based on fine-tuning, to formally describe the details of the personalization stage.
	The goal is to find a good model parameter for each client and we define $F:\mathbb{R}^{Nd1}\rightarrow\mathbb{R}$ as the global loss:
	
	\begin{equation}
		\min_{\theta_1,\theta_2,...,\theta_N\in \mathbb{R}^{d_1}}F(\theta_1,\theta_2,...,\theta_N) = \frac{1}{N}\sum_{i=1}^{N}f_i(\theta_i)
	\end{equation}
	
	\begin{equation}
		f_i(\theta_i) =\mathbb{E}_{(x,y)\sim\mathcal{D}_i}[L_i(\widehat{y}, y)]\label{perft:loss}
	\end{equation}
	
	\begin{equation}
		\widehat{y}=M_G(\theta_i;x)
	\end{equation}
	where the $\theta_i$ is what we want for the client $i$ through PFL, and the $ \widehat{y} $ is the output pseudo label corresponding to x.
	
	In PFL-FT, the initial value of $\theta_i$ is the global model parameter $\theta$, and the $\theta_i$ is updated by stochastic gradient descent (SGD) with a relatively small learning rate:
	\begin{equation}
		\theta_i = \theta_i - \alpha \cdot \triangledown{f_i(\theta_i)}
	\end{equation}
	where $\alpha$ is the learning rate of the $\theta_i$.
	
	\subsection{Mixing}
	The proposed PFL-MoE mixes the output of the personalized model and the global model using MoE architecture. MoE architecture is a method of ensemble learning. Different from the typical ensemble learning methods, the individual learners of MoE are trained for different subtasks, so its diversity can be guaranteed. In the MoE architecture, experts are independent on each other, and there is no limit that each expert must have the same mode structure \cite{peterson2019private}. 
	
	\begin{algorithm}[htbp]
		\renewcommand{\algorithmicrequire}{\textbf{Input:}}
		\renewcommand{\algorithmicensure}{\textbf{Output:}}
		\caption{PFL-MoE}
		\label{alg:pfl-moe}
		\begin{algorithmic}[1]
			
			\REQUIRE global model structure $M_G$,  and latest global model parameter $\theta$, client's data $D_i$, $\theta_i$'s learning rate $\alpha$, $w_i$'s learning rate $\beta$, local adaptation epochs $E$, loss function $L$
			
			\ENSURE personalized model parameter $\theta_i$, gating network parameter $w_i$
			\STATE \textit{\# Initialize client's personalized model parameter $\theta_i$ with global $\theta$}
			\STATE $\theta_i \leftarrow \theta$
			\STATE divide $D_i$ to $D_i^{per}$, $D_i^{gate}$
			\FOR{epoch $e = 1$ to $E$ }
			\STATE{\textit{\# Local adaptation}}
			\FOR{batch $(x, y) \subset D_i^{per} $ } 
			\STATE $\widehat{y} \leftarrow M_C(\theta_i; x)$
			\STATE $\theta_i \leftarrow \theta_i - \alpha \cdot \triangledown_{\theta_i}{L(\widehat{y}, y)}$ \textit{\# if using fine-tuning}
			\ENDFOR
			\STATE \textit{\# Mixing two experts}
			\FOR{batch $(x, y) \subset D_i^{gate}$}
			\STATE $g \leftarrow sigmoid(G(w_i; x))$
			\STATE $\tilde{y} \leftarrow g \cdot M_G(\theta; x) + (1-g) \cdot M_G(\theta_i; x) $
			\STATE $w_i \leftarrow w_i - \beta \cdot \triangledown_{w_i}{L(\tilde{y}, y)}$ 
			\ENDFOR
			\ENDFOR
			
		\end{algorithmic}
	\end{algorithm}
	
	More formally, in MoE architecture, the mxied output $\tilde{y} = \sum_{i=1}^{n}G(x)_i \cdot M_i(x)$, where $\sum_{i=1}^{n}G(x)_i=1$ and $M_i, i=1,...,n$ are $n$ experts (neural networks). Specifically, the gating network $G$ produces a probability distribution on $n$ experts, and the final output is the weighted sum of all experts.
	In our method, there are $n=2$ experts: the personalized model is viewed as the local expert, while the global model is viewed as the global expert. Their outputs are mixed by the gating network. The gating is a linear neural network, and can also be trained with SGD. 
	
	We define the gating network as $G$. For client $i$, we define its parameter of $G$ as $w_i$ ($d_2$-dimensional vector), and the loss function of gating $p_i$ is given by Eq. \ref{gate:mixloss}:
	\begin{equation}
		p_i(w_i) =\mathbb{E}_{(x,y)\sim\mathcal{D}_i}[L_i(\tilde{y}, y)] \label{gate:mixloss}
	\end{equation}
	where $p_i: \mathbb{R}^{d_2}\rightarrow\mathbb{R}$, and the mixed pseudo label $\tilde{y}$ is given by Eq. \ref{per:gateout} and Eq. \ref{per:mixture} when the input is $x$:
	\begin{equation}
		g = sigmoid(G(w_i; x))  \label{per:gateout}
	\end{equation}
	
	\begin{equation}
		\tilde{y}=g \cdot M_G(\theta;x) + (1-g) \cdot M_G(\theta_i;x) \label{per:mixture} 	
	\end{equation}
	where the $G$ is a linear neural network. $sigmoid $ is the activation function of the model output, and the term $g$ represents the mixing ratio of the expert models.
	For details, the number of the input neurons of $G$ is equal to the dimension of $x$, while the number of the output neurons of $G$ is equal to the number of experts. Particularly, similar to Peterson et al. \cite{peterson2019private}, we set the number of output neurons of $G$ to 1, so that $g$ is a 1-dimensional vector. At this setting, $g$ is the weight of the global expert, while $1-g$ is that of local expert. For the client $i$, to minimize $p_i(w_i)$, the gating parameter $w_i$ can be updated using SGD:
	\begin{equation}
		w_i = w_i - \beta \cdot \triangledown{p_i(w_i)} \label{gate:update}
	\end{equation}
	where the term $\beta$ refers to the learning rate of $w_i$. Alg.\ref{alg:pfl-moe} shows the details of the PFL-MoE.

	\section{PFL-MF and PFL-MFE} \label{sec:pflmf}
	PFL-FB \cite{yu2020salvaging} is a local adaption PFL algorithm based on freeze-base that freezes the base layer of the federated model and fine-tunes only the top layers. PFL-FB has the most significant personalization effect on the comparison of several methods for adapting federated global model to individual client.
	
	Following the three steps of PFL-MoE, we propose the PFL-MF algorithm by using the local expert generated by the PFL-FB algorithm.
	PFL-FB is a variant of PFL-FT \cite{wang2019federated}, where the global model structure $M_ G$ is divided into two parts: feature extractor $M_ E$ and classifier $M_C$.
	We define $\theta_E$ and $\theta_C$ as the parameter of $M_E$ and $M_C$ respectively. In PFL-MF, assume that the $\theta$ is the parameter of global model, for client $i$, $(\theta_E, \theta_{C_i})$ is initialized with $\theta$, where the term $\theta_{C_i}$ is the personalized classifier parameter of the client and the $\theta_E$ is shared to all clients.
	In personalization stage, the $\theta_{C_i}$ is updated by:

	\begin{equation}
		\theta_{C_i} = \theta_{C_i} - \alpha \cdot \triangledown{f_i(\theta_{E},\theta_{C_i})}
	\end{equation}
	
	\begin{equation}
		f_i(\theta_E,\theta_{C_i}) =\mathbb{E}_{(x,y)\sim\mathcal{D}_i}[L_i(\widehat{y}, y)] \label{perfb:loss}
	\end{equation}
	where the pseudo label $\widehat{y}$ corresponding to x is given by:
	\begin{equation}
		a = M_E(\theta_E, x)
	\end{equation}
	
	\begin{equation}
		\widehat{y}=M_C(\theta_C; a)
	\end{equation}
	where $a$ is the activations between $M_E$ and $M_C$.
	
	In the mixing stage, $w_i$ is updated by Eq. \ref{gate:update}, and the gating loss $p_i(w_i)$ is defined in Eq. \ref{gate:mixloss}. In the PFL-MF algorithm, different from Eq. \ref{per:mixture}, the mixed pseudo label $\tilde{y}$ corresponding to $x$ is written as:
	
	\begin{equation}
		\tilde{y}=g \cdot M_C(\theta_C; a) + (1-g) \cdot M_C(\theta_{C_i}; a)
	\end{equation}
	
	Through experiments on the Fashion-MNIST dataset, we observe that PFL-MF can achieve good personalization and maintain good generalization capability.
	However, due to the limitation of the simple linear network, the gating works not well when the input is more complex high-dimensional data. 
	
	To address this issue, we propose PFL-MFE algorithm that using features as the input of gating. 
	In convolutional neural networks, it is recognized that the base layer is responsible for capturing general features, while the deeper part is responsible for classification. 
	Intuitively, the features are a better choice for the gating input compared to raw data. Based on this idea, in the PFL-MFE algorithm, we modify the model structure of the PFL-MF algorithm, and use the activations $a$ as the gating input:
	\begin{equation}
		g = sigmoid(G(w_i; a))
	\end{equation}
	at this time, the number of the input neurons of $G$ is equal to the dimension of $a$ instead of $x$.
	Alg. \ref{alg:pfl-mf} describes the implementation details of the  PFL-MF and PFL-MFE algorithms for client $i$. 
	\begin{algorithm}[htbp]
		\renewcommand{\algorithmicrequire}{\textbf{Input:}}
		\renewcommand{\algorithmicensure}{\textbf{Output:}}
		\caption{PFL-MF(E) Algorithm}
		\label{alg:pfl-mf}
		\begin{algorithmic}[1]
			\REQUIRE global extractor model structure $M_E$ and its parameter $\theta_E$, global classifier model structure $M_C$ and its parameter $\theta_C$, client's data $D_i$, $\theta_{C_i}$'s learning rate $\alpha$, $w_i$'s learning rate $\beta$, local adaptation epochs $E$, loss function $L$, if variant flag $v$
			
			\ENSURE personalized classifier parameter $\theta_{C_i}$, gating network parameter $w_i$
			\STATE \textit{\# Initialize client's personalized classifier parameter $\theta_{C_i}$ with global $\theta_C$}
			\STATE $\theta_{C_i} \leftarrow \theta_C$
			\STATE divide $D_i$ to $D_i^{per}$, $D_i^{gate}$
			\FOR{epoch $e = 1$ to $E$ }
			\STATE{\textit{\# Local adaptation}}
			\FOR{batch $(x, y) \subset D_i^{per} $ } 
			\STATE $\widehat{y} \leftarrow M_C(\theta_C; M_E(\theta_E; x))$
			\STATE $\theta_{C_i} \leftarrow \theta_{C_i} - \alpha \cdot \triangledown_{\theta_{C_i}}{L(\widehat{y}, y)}$
			\ENDFOR
			\STATE \textit{\# Mixing two experts}
			\FOR{batch $(x, y) \subset D_i^{gate} $ }
			\STATE $a \leftarrow M_E(\theta_E; x)$
			\IF{ $v$ }
			\STATE $g \leftarrow sigmoid(G(w_i; a))$ \textit{\# PFL-MFE} 
			\ELSE 
			\STATE $g \leftarrow sigmoid(G(w_i; x))$ \textit{\# PFL-MF}
			\ENDIF
			\STATE $\tilde{y} \leftarrow g \cdot M_C(\theta_C; a) + (1-g) \cdot M_C(\theta_{C_i}; a) $
			\STATE $w_i \leftarrow w_i - \beta \cdot \triangledown_{w_i}{L(\tilde{y}, y)}$ 
			\ENDFOR
			\ENDFOR
			
		\end{algorithmic}
	\end{algorithm}

	\begin{figure*}[htbp]
		\centerline{\includegraphics[width=1\textwidth]{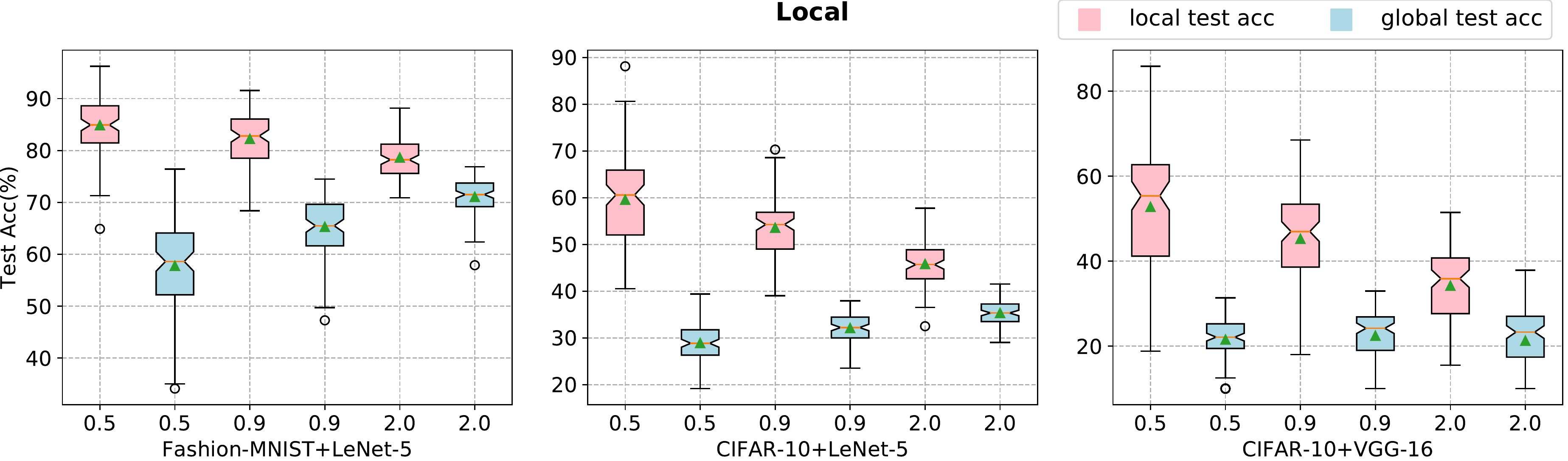}}
		\caption{The local test accuracy and global test accuracy in Local for 100 clients. Each box contains 100 test accuracy of 100 clients for each specific $\alpha$. 
		}
		\label{fig:local_acc}
	\end{figure*}

	\section{Experimental Evaluation}\label{sec:dbe}
	
	\subsection{Experimental Setup}
	
	\subsubsection{Datasets and models}
	In our experiments\footnote{All code and results can be found at \url{https://github.com/guobbin/PFL-MoE}.}, we use two image recognition datasets to
	conduct model training: Fashion-MNIST \cite{xiao2017/online} and CIFAR-10 \cite{krizhevsky2009learning}. With two network models trained, we have three combinations: Fashion-MNIST + LeNet-5 \cite{lecun1998gradient}, CIFAR-10 + LeNet-5, and CIFAR-10 + VGG-16 \cite{2014Very}. 
	Fashion-MNIST is an alternative dataset of MNIST. It contains 28$\times$28 grayscale images of the front side of 70,000 different products in 10 categories, divided into 60000/10000 training/testing datasets. CIFAR-10 is a dataset of $32\times32$ color images closer to universal objects, with 3 channels and a training/testing dataset division of 50000/10000. LeNet-5 contains two convolutional layers and three fully connected layers. We resize the images of the datasets to $32\times32$  uniformly, and due to the difference in the number of image channels, the corresponding number of the LeNet-5 model parameters are 61706 and 62006, respectively. VGG-16 is a relatively deep neural network, including 13 convolutional layers and 3 fully connected layers, with a total of 15,245,130 parameters. 
	
	\subsubsection{Non-IID Data}
	Similar to Hsu et al. \cite{49350}, we use Dirichlet distribution with $\alpha$ to simulate the non-IID data distributions. $\alpha>0$ is a concentration parameter controlling the identicalness among clients. \textbf{The larger the $\alpha$, the higher the similarity between the distributions}, and we chose $\alpha = [0.5, 0.9, 2] $ to generate three non-IID data distributions for experiments. For each $\alpha$, we split the training set and allocate images from each class to 100 clients.
	
	\subsubsection{Testing}
	Each client has two types of tests, including local test and global test. Local test uses the test set of the same distribution as the training set of the same client. In contrast, global test uses the global test set (10000 test images).
	Following the local test method of Yu et al. \cite{yu2020salvaging}, for each client, we calculate the client's per-class accuracy on the global test set, multiply it by the corresponding class’s ratio in the training set of the client, and sum up the resulting values. The accuracy of the local test and global test is used as a measure of personalization and generalization.
	\subsubsection{Baselines and PFL-MoE}
	To verify the effectiveness of the proposed personalization algorithms PFL-MF and PFL-MFE, we compare them with three baselines: Local, FedAvg, and PFL-FB. 
	The detailed settings of each algorithm are as follows:
	\begin{table*}[htbp]
		\centering
		\caption{The average value of \textbf{local test} accuracy of all clients in three baselines and proposed algorithms. Bold means the best in all methods. 
		}
		\begin{tabular}{ccccccc}
			\toprule
			& non-IID $\alpha$  & Local(\%) & FedAvg(\%) & PFL-FB(\%) & PFL-MF(\%) & PFL-MFE(\%) \\
			\midrule
			\multicolumn{1}{c}{\multirow{3}[0]{*}{\tabincell{c}{Fashion-MNIST \& \\ LeNet5}}} & 0.5   & 84.87 & 90    & 92.84 & 92.85 & \textbf{92.89} \\
			& 0.9   & 82.23 & 90.31 & 91.84 & \textbf{92.02} & 92.01 \\
			& 2     & 78.63 & 90.5  & 90.47 & \textbf{90.97} & 90.93 \\
			\midrule
			\multicolumn{1}{c}{\multirow{3}[0]{*}{\tabincell{c}{CIFAR-10 \& \\ LeNet5}}} & 0.5   & 65.58 & 68.92 & \textbf{77.46} & 75.49 & 77.23 \\
			& 0.9   & 61.49 & 70.7  & 74.7  & 74.1  & \textbf{74.74} \\
			& 2     & 55.8  & 72.69 & 72.5  & 73.24 & \textbf{73.44} \\
			\midrule
			\multicolumn{1}{c}{\multirow{3}[0]{*}{\tabincell{c}{CIFAR-10 \& \\ VGG-16}}} & 0.5   & 52.77 & 88.16 & \textbf{91.92} & 90.63 & 91.71 \\
			& 0.9   & 45.24 & 88.45 & \textbf{91.34} & 90.63 & 91.18 \\
			& 2     & 34.2  & 89.17 & \textbf{90.4} & 90.15 & \textbf{90.4} \\
			
			\bottomrule	
		\end{tabular}%
		\label{tab:local acc}%
	\end{table*}%
	
	\begin{table*}[htbp]
		\centering
		\caption{The average value of \textbf{global test} accuracy of all clients. Bold means the best in all personalization algorithms. For FedAvg, all clients have the same global model, and they use the same global test set for the global test, so their global test accuracy is the same. The value is exactly the average local test accuracy of above local test, due to the sum of the proportions of each category of all clients is the same as the global test set.\\}
		
		\begin{tabular}{ccccccc}
			\toprule
			& non-IID $\alpha$  & Local(\%) & FedAvg(\%) & PFL-FB(\%) & PFL-MF(\%) & PFL-MFE(\%) \\
			\midrule
			\multicolumn{1}{c}{\multirow{3}[0]{*}{\tabincell{c}{Fashion-MNIST \& \\ LeNet5}}} & 0.5   & 57.77 & 90    & 83.35 & \textbf{85.45} & 85.3 \\
			& 0.9   & 65.28 & 90.31 & 85.91 & \textbf{87.69} & 87.67 \\
			& 2     & 71.06 & 90.5  & 87.77 & \textbf{89.37} & 89.18 \\		
			\midrule
			\multicolumn{1}{c}{\multirow{3}[0]{*}{\tabincell{c}{CIFAR-10 \& \\ LeNet5}}} & 0.5   & 28.89 & 68.92 & 54.28 & \textbf{62.33} & 58.27 \\
			& 0.9   & 32.1  & 70.7  & 59.93 & \textbf{65.78} & 64.13 \\
			& 2     & 35.32 & 72.69 & 66.06 & \textbf{69.79} & 69.78 \\
			\midrule
			\multicolumn{1}{c}{\multirow{3}[0]{*}{\tabincell{c}{CIFAR-10 \& \\ VGG-16}}} & 0.5   & 21.53 & 88.16 & 82.39 & \textbf{85.81} & 84.05 \\
			& 0.9   & 22.45 & 88.45 & 82.62 & \textbf{88.15} & 87.9 \\
			& 2     & 21.27 & 89.17 & 88.77 & \textbf{89.3} & \textbf{89.3} \\
			\bottomrule	
		\end{tabular}%
		\label{tab:global acc}%
	\end{table*}%

	\paragraph{Local}
	Each client's local model was trained with its local training set for 300 epochs with the regularization factor $\lambda=0.0005$, batch size $B=64$. The optimizer is SGD with the momentum $m=0.9$, learning rate $\eta=0.1$, and learning rate decay $\gamma=0.1$ every 100 epochs or Adam. Each Local experiment chose the better one as the final experimental result. The results of the Local are shown in Fig. \ref{fig:local_acc}.
	
\begin{figure}[htbp]
		\centerline{\includegraphics[width=0.5\textwidth]{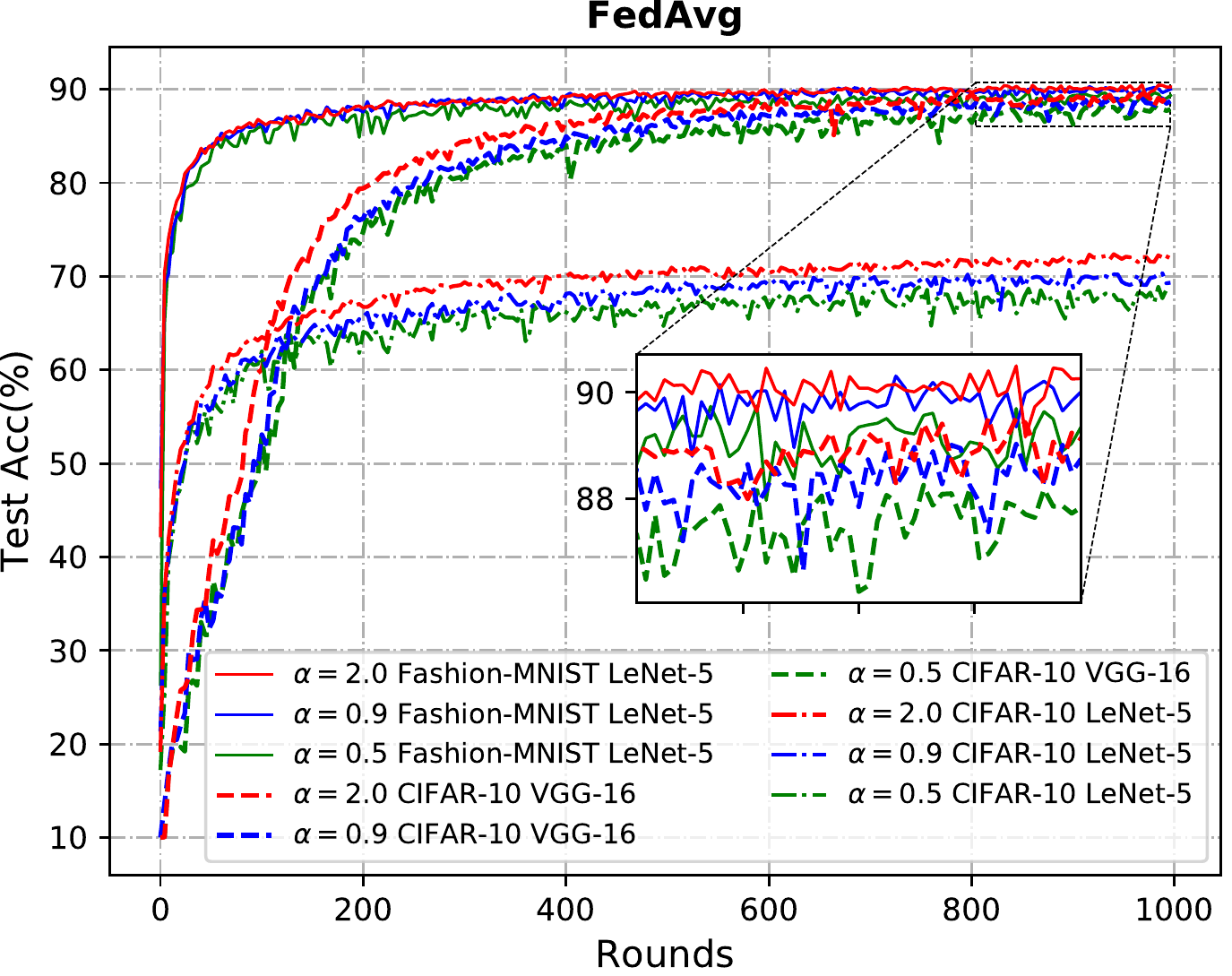}}
		\caption{Global test accuracy of FedAvg, 1000 rounds, participate rate 0.1, 100 clients.}
		\label{fig:fed_acc}
	\end{figure}

	\paragraph{FedAvg}
	We trained the global model for 1000 rounds, following McMahan et al. \cite{mcmahan2017communication}, each round randomly selected $c=0.1$ rate of the clients for training with the learning rate $\eta=0.01$, momentum $m=0.5$, local batch size $b=10$, and local epoch $e=5$. We saved the model parameters that performed best on the global test set in these 1,000 rounds for the local adaptation of the following personalization algorithms. The results of FedAvg algorithm are shown in Fig.\ref{fig:fed_acc}.
	
	\paragraph{Personalization algorithms}
	Each client trained 200 epochs for local adaption using the same SGD optimizer as the Local baseline. In PFL-FB experiments, the convolutional layers of LeNet-5 and VGG-16 were fixed, and the top layers were trained with the learning rate $\alpha=0.001$ and $\alpha=0.005$, respectively. In PFL-MF experiments, the optimizer of gating network was SGD with the learning rate $\beta=0.001$. The input layer of gating network has $32\times32\times1=1024$ and $32\times32\times3=3072$ neurons for Fashion-MNIST and CIFAR-10, respectively. But in PFL-MFE, it changes to 400 and 512 for LeNet-5 and VGG-16.
	\subsection{Effect of non-IID}
	
	Fig. \ref{fig:local_acc} shows that \textbf{as the degree of non-IID increases ($\alpha$ get smaller), the local test accuracy of the Local also increases.} It consistent with our intuition: the more skewed the client's data distribution, the easier it is to get high local test accuracy (e.g., in a client, a few categories account for a large proportion while only a small amount of data belongs to other categories, or the client has only one category at all). However, The model in Local is not necessarily a good model because the categories with only a small amount of data may be discarded, so that it never really distinguishes the difference between each category. The conjecture is confirmed from the global test of the Local in Fig. \ref{fig:local_acc} and Tab. \ref{tab:global acc}., in all experiment, the global test accuracy of all clients is particularly low.
	The highly non-IID skewed local data leads to the limitation of the stand-alone trained model.
	
	From Tab. \ref{tab:local acc} and Tab. \ref{tab:global acc}, in all experiments, \textbf{the accuracy of Local is lower than FedAvg's in both local test and global test}, whether it is a large model VGG-16 or a small model LeNet-5. 
	In the CIFAR-10 experiments of Local,
	\textbf{the accuracy of the larger model VGG-16 is particularly low, which is significantly lower than the smaller model LeNet-5}. Generally speaking, the deeper the model, the higher the limitation of its capabilities, but more data is required for training. 
	Although the simple model performs relatively well in Local, its limitation gives the client a stronger incentive to participate in federated learning to train a more powerful model. Fig. \ref{fig:fed_acc} shows that VGG-16 achieves nearly 90\% test accuracy by FedAvg algorithm. Large model can be adequately trained through federated learning, which greatly improves model accuracy compared to Local.
	
	The column FedAvg in Tab. \ref{tab:local acc} (1000th round in Fig. \ref{fig:fed_acc}) presents the test accuracy of the global model of the FedAvg, where all three groups of experiments have the same results: \textbf{the smaller the $\alpha$, the worse the performance of FedAvg.} 
	On the contrary, Tab. \ref{tab:local acc} shows that \textbf{as $\alpha$ gets smaller, the potential for personalization is growing}.
	In DML, the convergence relies on the assumption that data are IID among workers. In line with the theoretical analysis \cite{DBLP:conf/iclr/LiHYWZ20}, a smaller $\alpha$ means more deviation from the IID assumption, which hurts convergence rate and final model performance of FedAvg.
	But this also makes us think: \textit{Why use the same global model for clients with heterogenous data.}
	We further show that federated personalized learning is naturally suitable for the non-IID data distribution, and can effectively improve overall performance. 
	
	\begin{figure}[htbp]
		\centerline{\includegraphics[width=0.5\textwidth]{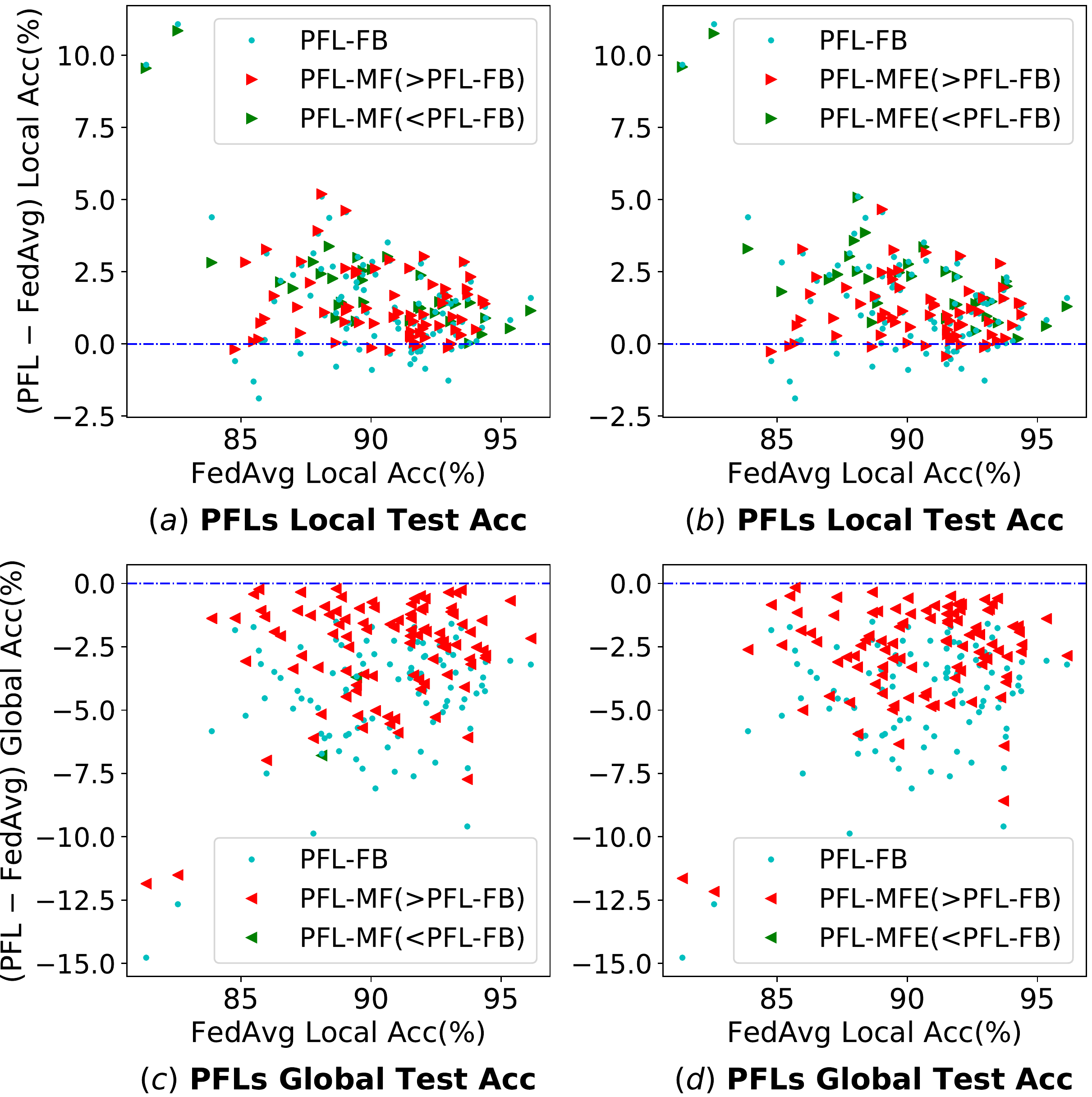}}
		\caption{Fashion-MNIST + LeNet-5, $\alpha=0.9$, 100 clients. The global test accuracy and local test accuracy of all client of PFL-FB, PFL-MF, and PFL-MFE algorithms. All x-axis are FedAvg local test accuracy of each client (can be regarded as client index). Each point represents a test accuracy comparison between a PFL algorithm and FedAvg for a particular client. \textbf{Compared with PFL-FB baseline, the more red triangles, the better our algorithms.} Fig. \ref{fig:fmnist_lenet}$(a)$ and Fig. \ref{fig:fmnist_lenet}$(b)$ compare the local test accuracy of the four algorithms for each client, while Fig. \ref{fig:fmnist_lenet}$(c)$ and Fig \ref{fig:fmnist_lenet}$(d)$ compare the global test accuracy of them for each client. 
		Fig.\ref{fig:cifar_lenet} and Fig.\ref{fig:cifar_lenet} have the same meaning but with different dataset+model. 
		}
		\label{fig:fmnist_lenet}
	\end{figure}
	
	\subsection{Personalization of PFL-FB and PFL-MF}
	Tab. \ref{tab:global acc} shows that the average global test accuracy of these three personalization algorithms are higher than that of Local, which means that personalized federated learning can effectively improve the generalization ability of the model on client. 
	Tab. \ref{fig:local_acc} shows that the average local test accuracy of these three personalization algorithms are 2\% $\sim$ 10\% higher than that of FedAvg at $\alpha=0.5$.
	However, \textbf{PFL-FB is more likely to cause overfitting}. When $\alpha$ is relatively large,
	PFL-FB is more likely to overfit local data, which makes the personalized model worse than the global model. In the proposed algorithm PFL-MF, if the personalized model deteriorates, it will be directly discarded by the gating network. Moreover, the gating network has learned which data is appropriate to use the personalized model and which data needs to be weighted more toward the global model. Consequently, PFL-MF is more stable, and its local test accuracy is always higher than FedAvg. 
	
	Comparing PFL-FB with PFL-MF in Tab. \ref{tab:local acc} and Tab. \ref{tab:global acc}, for each client,
	\textbf{the increase of the local test accuracy of PFL-FB comes at the cost of a significant drop in global test accuracy.}
	For example, in the CIFAR-10 + LeNet-5 experiment, the average global test accuracy of PFL-FB dropped by 15\%, from 68.92\% to 54.28\%. PFL-MF achieves the average global test accuracy with 1\% $\sim$ 8\% higher than that of PFL-FB in all experiments. 
	The main incentive for clients participating in federated learning is to obtain a better model that is both personalized and generalized. 
	However, for a single model, the two capabilities are often contradictory. The algorithm based on our PFL-MoE makes it possible to get both capabilities at the same time.
	
	As shown in Tab. \ref{tab:local acc} and Tab. \ref{tab:global acc}, in experiments on Fashion-MNIST, the proposed PFL-MF algorithm performs better than PFL-FB on the two tests, which indicates that
	\textbf{PFL-MF achieves better personalization and better generalization than PFL-FB.} 
	However, in experiments on CIFAR-10, the average local test accuracy of PFL-MF is lower than that of PFL-FB, except in the LeNet-5 experiment with $\alpha=2$. It can be observed more obviously from Fig. \ref{fig:fmnist_lenet}$(a)$, Fig. \ref{fig:cifar_lenet}$(a)$, and Fig. \ref{fig:cifar_vgg}$(a)$. In Fig. \ref{fig:fmnist_lenet}$(a)$, red is the majority, while in Fig. \ref{fig:cifar_lenet}$(a)$ and Fig. \ref{fig:cifar_vgg}$(a)$, green is the majority.
	In PFL-FM, the gating network can only recognize and make the decisions for simple data, such as the Fashion-MNIST dataset. But for the more complex color image dataset CIFAR-10, the gating network needs to be improved.
	`
	\subsection{Effect of PFL-MFE}
\begin{figure}[htbp]
		\centering
		\centerline{\includegraphics[width=0.5\textwidth]{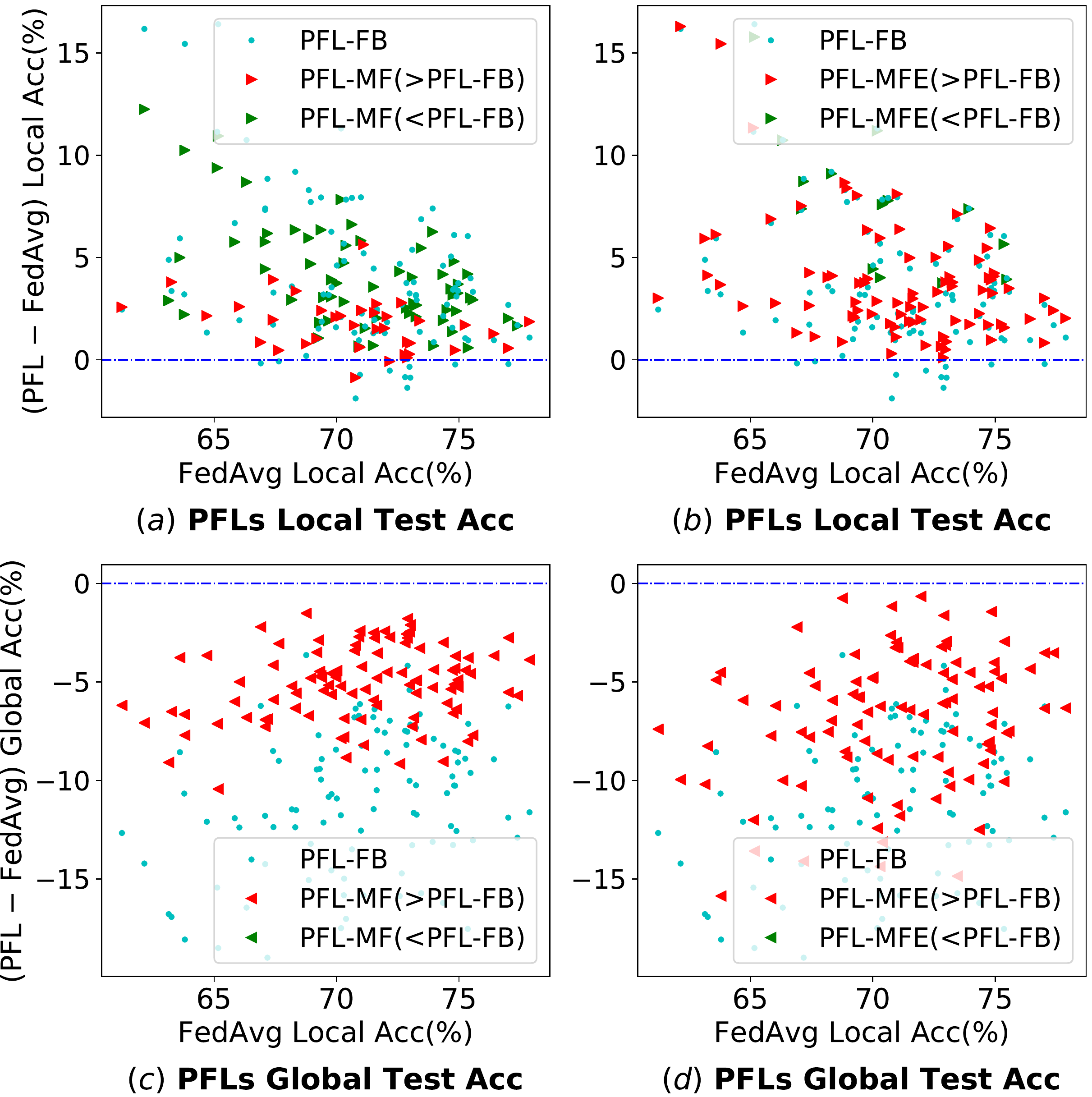}}
		\caption{CIFAR-10 + LeNet-5, $\alpha=0.9$, 100 clients. Comparing Fig. \ref{fig:cifar_lenet}$(a)$ with Fig. \ref{fig:cifar_lenet}$(b)$, the same in Fig. \ref{fig:fmnist_lenet}, the points with $y>0$ in the triangle points are more than that in round points. But in Fig. \ref{fig:cifar_lenet}$(a)$, there are a large number of green triangles which means that the personalization effect of PFL-MF is not better as that of PFL-FB. But from Fig. \ref{fig:cifar_lenet}$(b)$, the extension algorithm FL-MFE  works better than PFL-FB.}
		\label{fig:cifar_lenet}
	\end{figure}

\begin{figure}[htbp]
		\centering
		\centerline{\includegraphics[width=0.5\textwidth]{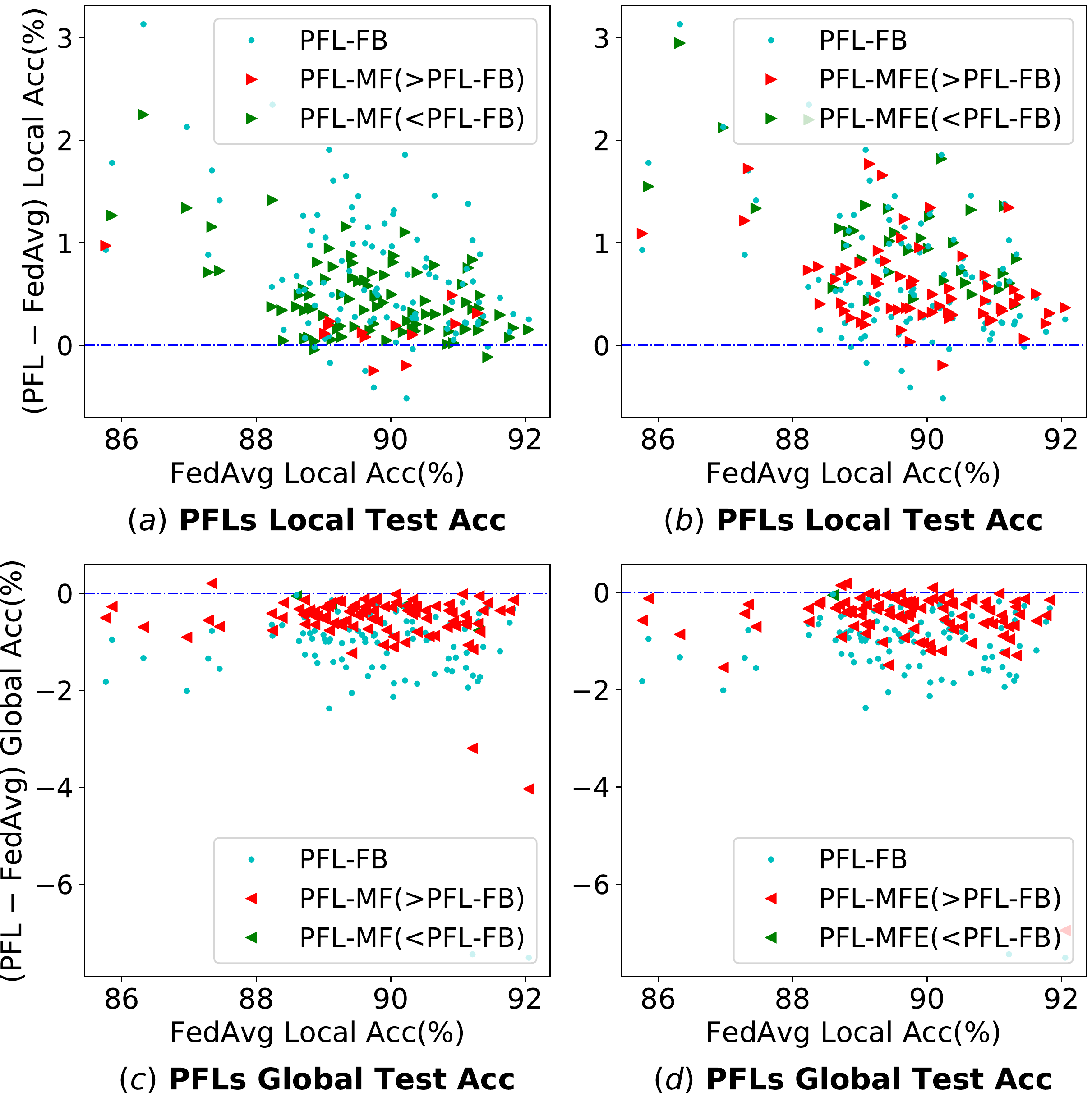}}
		\caption{CIFAR-10 + VGG-16, $\alpha=2.0$, 100 clients. Just like Fig. \ref{fig:cifar_lenet}, PFL-MFE shows its advantage on the CIFAR-10 dataset. Comparing Fig. \ref{fig:cifar_vgg}$(c)$ with Fig. \ref{fig:cifar_vgg}$(d)$, almost all triangles are red, which is also shown in Fig. \ref{fig:fmnist_lenet} and Fig. \ref{fig:cifar_lenet}, indicating that our algorithms are always superior to PFL-FB in the global test.} 
		\label{fig:cifar_vgg}	
	\end{figure}

	From Tab. \ref{tab:local acc}, in the experiments on CIFAR-10, the average local test of PFL-MFE is 0.3\% $\sim$ 1.7\% higher than that of PFL-MF.
	Although the average global test accuracy of PFL-MFE is slightly lower than that of PFL-MF, it is still higher than that of PFL-FB.
	Comparing Fig. \ref{fig:cifar_lenet}$(a)$ with Fig. \ref{fig:cifar_lenet}$(b)$, we can see that, lots of green triangles in Fig. \ref{fig:cifar_lenet}$(a)$ increase and turn to red in Fig. \ref{fig:cifar_lenet}$(b)$ while the triangles in Fig. \ref{fig:cifar_lenet}$(c)$ and Fig. \ref{fig:cifar_lenet}$(d)$ are always red. In Fig. \ref{fig:cifar_vgg}, we have the same observation as in Fig. \ref{fig:cifar_lenet}.
	In the experiments on Fashion-MNIST, although PFL-MFE outperforms PFL-MF in local test only when $\alpha=0.5$, it is very close to PFL-MF at other $\alpha$ values.
	PFL-MFE and PFL-MF  are both 0.05\% $\sim$ 0.5\% higher than PFL-FB in average local test accuracy. 
	
	In the experiments on CIFAR-10, PFL-MFE is more effective than PFL-MF, because the gating network can make better decisions indirectly by using the abstract features. For the simple Fashion-MNIST dataset, PFL-MF is enough. 
	\textbf{PFL-MFE can better recognize complex input data while maintaining the advantages of PFL-MF.}
	As shown in Tab. \ref{tab:local acc} and Tab. \ref{tab:global acc}, in all experiments, overall, PFL-MFE achieves the best average local test accuracy among three personalization algorithms, and it is much better than PFL-FB in the average global test accuracy. In a word, the proposed PFL-MFE algorithm can realize personalization more effectively while maintaining better generalization.
	
	\section{Conclusion and Future work}\label{sec:conclusion}
	We propose PFL-MoE, an MoE based approach for personalized federated learning. PFL-MoE can combine advantages of the personalized model and the global model and achieve both better personalization and better generalization.
	PFL-MF, an instance of PFL-MoE, performs better than existing algorithm PFL-FB in both local test and global test. We also show that PFL-MFE strengthens the decision-making ability of the MoE gating network and can be effective when processing more complex data.
	In the future, we will explore the instance of PFL-MoE of other personalized federated learning algorithms and compare their overall performance.
	
	\bibliographystyle{./bibliography/IEEEtran}
	\bibliography{./bibliography/IEEEabrv,./bibliography/IEEEexample}
	
\end{document}